%% file: naaclhlt2019.tex
\documentclass[11pt]{article}
\usepackage[hyperref]{naaclhlt2019}
\usepackage{times}
\usepackage{latexsym}
\usepackage{latexsym}
\usepackage{booktabs}
\usepackage{multirow}
\usepackage{pgfplotstable}
\usepackage{filecontents}
\usepackage{array}
\usepackage{xcolor,colortbl}
\usepackage{ctable}
\usepackage{float}
\usepackage{subcaption}
\usepackage{array}
\usepackage{nicefrac}
\usepackage{amsmath}
\usepackage[bottom]{footmisc}
\usepackage{algorithm2e}
\usepackage[bottom]{footmisc}

\usepackage{pgf}
\usepackage{graphicx}
\usepackage{tikz}
\usetikzlibrary{shapes}
\usetikzlibrary{arrows}
\usetikzlibrary{positioning}

\usepackage{pgfplots}
\usepackage{makecell}
\usepackage{layouts}
\usepackage[margin=0.1cm]{caption}
\newcolumntype{?}{!{\vrule width 1.5pt}}

\newrobustcmd{\B}{\bfseries}

\usepackage{array}
\newcolumntype{x}[1]{>{\centering\let\newline\\\arraybackslash\hspace{0pt}}p{#1}}
\newcolumntype{?}{!{\vrule width 1.5pt}}

\usepackage{url}

\aclfinalcopy 

\title{Scalable Cross-Lingual Transfer of Neural Sentence Embeddings}

\author{Hanan Aldarmaki$^1$ \and Mona Diab$^{1,2}$\\
    $^1$The George Washington University\\%
    $^2$AWS, Amazon AI\\
    {\tt aldarmaki@gwu.edu,diabmona@amazon.com}
}

\date{}

\begin{document}
\maketitle
\begin{abstract}

We develop and investigate several cross-lingual alignment approaches for neural sentence embedding models, such as the supervised inference classifier, InferSent, and sequential encoder-decoder models. We evaluate three alignment frameworks applied to these models:  joint modeling, representation transfer learning, and sentence mapping, using parallel text to guide the alignment. Our results support representation transfer as a scalable approach for modular cross-lingual alignment of neural sentence embeddings, where we observe better performance compared to joint models in intrinsic and extrinsic evaluations, particularly with smaller sets of parallel data. 

\end{abstract}

\section{Introduction}

Probabilistic sentence representation models generally fall into two categories: bottom-up compositional models, where sentence embeddings are composed from word embeddings via a linear function like averaging, and top-down compositional models that are trained with a sentence-level objective, typically within a neural architecture. Sequential data like sentences can be modeled using recurrent, recursive, or convolutional networks, which can implicitly learn intermediate sentence representations suitable for each learning task. Depending on the training objective, these intermediate representations sometimes encode enough semantic and syntactic features to be suitable as general-purpose sentence embeddings. For examples, it was shown in \newcite{conneau2017supervised} that a model trained to maximize inference classification accuracy can yield generic representations that perform well across a wide set of extrinsic classification benchmarks. Other training objectives, like denoising auto-encoders or neural sequence to sequence models \cite{hill2016learning}, can also yield general-purpose representations with different characteristics. While bottom-up models can achieve superior performance in tasks that are independent of syntax, such as topic categorization, neural models often yield representations that encode syntactic and positional features, which results in superior performance in tasks that rely on sentence structure \cite{aldarmaki2018evaluation}.

General-purpose sentence embeddings can be used as features in various classification tasks, or to directly assess the similarity of a pair of sentences using the cosine measure. It is often desired to generalize word and sentence embeddings across several languages to facilitate cross-lingual transfer learning \cite{zhou2016cross} and mining of parallel sentences \cite{guo2018effective}. For word embeddings, cross-lingual learning can be achieved in various ways \cite{upadhyay2016cross}, such as learning directly with a cross-lingual objective \cite{glove-15}  or post-hoc alignment of monolingual word embeddings using dictionaries  \cite{AmmarMTLDS16}, parallel corpora \cite{gouws2015bilbowa,Klement-12}, or even with no bilingual supervision \cite{conneau2017word,aldarmaki2018unsupervised}. For bottom-up composition like vector averaging, word-level alignment is sufficient to yield cross-lingual sentence embeddings. For top-down sentence embeddings, the efforts in cross-lingual learning are more limited. Typically, a multi-faceted cross-lingual learning objective is used to align the sentence models while training them, as in \newcite{soyer2014leveraging}. Cross-lingual sentence embeddings can also be learned via a neural machine translation framework trained jointly for multiple languages \cite{schwenk2017learning}. 

While they indeed yield cross-lingual embeddings, the joint training models in existing literature pose some practical limitations: simultaneous training requires massive computational  resources, particularly for sequential models like the bi-directional LSTM networks typically used to encode sentences. In addition, the joint framework does not allow post-hoc or modular training, where new languages can be added and aligned to existing pre-trained encoders. Recently, \newcite{conneau2018xnli} proposed an approach for cross-lingual sentence embeddings by aligning encoders of new languages to a pre-trained English encoder using parallel corpora. Such approach promises to be more suitable for modular training of general sentence encoders, although so far it has only been evaluated in natural language inference classification. 

In this paper, we develop and evaluate three alignment frameworks: joint modeling, representation transfer learning, and sentence mapping, applied on two modern general-purpose sentence embedding models: the inference-based encoder, InferSent \cite{conneau2017supervised}, and the sequential denoising auto-encoder, SDAE \cite{hill2016learning}. For most approaches, we rely on parallel sentences as sentence-level dictionaries for cross-lingual supervision. We report the performance  on sentence translation retrieval and cross-lingual document classification. Our results support representation transfer as a scalable approach for modular cross-lingual alignment that works well across different neural models and evaluation benchmarks. 

\section{Related Work}  

Learning bilingual compositional representations can be achieved by optimizing a bilingual objective on parallel corpora. In \newcite{Pham-15}, distributed representations for bilingual phrases and sentences are learned using an extended version of the paragraph vector model \cite{le2014distributed}  by forcing parallel sentences to share one vector. In \newcite{soyer2014leveraging}, cross-lingual compositional embeddings are learned by optimizing a joint bilingual objective that aligns parallel source and target representations by minimizing the Euclidean distances between them, and a monolingual objective that maximizes the similarity between similar phrases. The monolingual objective was implemented by maximizing the similarity between random phrases and subphrases within the same sentence. Cross-lingual representations can also be induced implicitly within a machine learning framework that is trained jointly for multiple language pairs. In \newcite{schwenk2017learning}, encoders and decoders for the given languages are trained jointly using a neural sequence to sequence model \cite{sutskever2014sequence} using parallel corpora that are partially aligned; that is, each language within a pair is also part of at least one other parallel corpus. Neural machine translation can also be achieved with a single encoder and decoder that handles several input languages \cite{johnson2016google}, but the latter has not been evaluated as a general-purpose sentence representation model. According to \newcite{hill2016learning}, the quality of the representations induced using a machine translation objective is lower than other neural models trained with different compositional objectives, such as Denoising Auto-Encoders and Skip-Thought \cite{kiros2015skip}. Mono-lingual evaluation of sentence representation models can be found in \newcite{hill2016learning}, \newcite{aldarmaki2018evaluation}, and \newcite{conneau2018senteval}. In \newcite{aldarmaki2016learning}, a modular training objective has been proposed for cross-lingual sentence embedding. However, their application was limited to the specific matrix factorization model they discussed. More recently, \newcite{conneau2018xnli} proposed a modular transfer learning objective and evaluated it on neural sentence encoders using cross-lingual natural language inference classification. Our representation transfer framework is very similar to their approach, although we use a simpler loss function. In addition, we evaluate the framework as a general-purpose sentence encoder and compare it to other frameworks.  

\section{Approach}

We selected two modern general-purpose sentence embedding models, the Inference-based classification model (\texttt{InferSent}) described in \newcite{conneau2017supervised}, and the Sequential Denoising Auto-Encoder (\texttt{SDAE}) described in \newcite{hill2016learning}. Both are implemented using a bidirectional LSTM network as an encoder followed by a classification or decoding network. We describe three possible cross-lingual alignment frameworks:

\paragraph{Joint cross-lingual modeling:} We extend the monolingual objective of each model to multiple languages to be trained simultaneously via direct cross-lingual interactions in the objective function. This is in line with most existing cross-lingual extensions for top-down compositional models 
\paragraph{Representation transfer learning:} We directly optimize the sentence embeddings of new languages to match their translations in a parallel language (i.e. English). A similar approach was independently developed in \newcite{conneau2018xnli}. 
\paragraph{Sentence mapping:} Following the modular alignment framework for word embeddings \cite{smith2017offline}, we fit an orthogonal transformation matrix on monolingual embeddings using a parallel corpus as a dictionary. Sentence mapping has been evaluated for word averaging models in \newcite{aldarmaki2019context}.  

\subsection{Architectures}

Most neural sentence embedding models are based on a sequential encoder---typically a bi-directional Long Short-Term Memory \cite{schuster1997bidirectional}---followed by either a sequential decoder or a classifier. These models can be categorized according to their training objective:

\paragraph{Classification Accuracy:} Sentence encoders can be trained by maximizing the accuracy in an extrinsic evaluation task. For example, \texttt{InferSent}  \cite{conneau2017supervised} is trained on the Stanford Natural Language Inference (SNLI) dataset for inference classification \cite{bowmanlarge}. This type of model requires labeled training data, which can make it challenging to expand across different languages. 

\paragraph{Reconstruction:} Using raw monolingual data, sentence encoders can be trained by minimizing the reconstruction loss, where a decoder is trained simultaneously to reconstruct the input sentence from the intermediate representation---e.g. Sequential Auto-Encoder (\texttt{SAE}) and Sequential Denoising Auto-Encoder (\texttt{SDAE}) \cite{hill2016learning}. The latter introduces textual noise on the input sentence to make the embeddings more robust. 

\paragraph{Translation:} In Neural Machine Translation (NMT), a model is trained to maximizes the accuracy of generating a translation from the intermediate representation of the source sentence. Unlike modern \texttt{NMT} systems that rely on attention mechanisms, this model is trained for the purpose of sentence embedding, so only the intermediate representations are used as input to the decoder. This model requires parallel corpora for training. 

\begin{figure}[ht]
\centering
\tikzset{every picture/.style={line width=0.75pt}}
\begin{subfigure}[b]{0.48\textwidth} 
\scalebox{0.8}{  
\hspace{-1cm}
\hspace{10pt}
\begin{tikzpicture}[x=0.75pt,y=0.75pt,yscale=-1,xscale=1]
\draw   (174.96,143.8) -- (246.96,143.8) -- (246.96,162.27) -- (174.96,162.27) -- cycle ;
\draw    (230.02,143.27) -- (230.02,162.27) ;
\draw    (211.86,143.27) -- (211.86,162.27) ;
\draw    (193.11,143.27) -- (193.11,162.27) ;
\draw    (211.96,181.27) -- (211.96,165.27) ;
\draw [shift={(211.96,163.27)}, rotate = 450] [color={rgb, 255:red, 0; green, 0; blue, 0 }  ][line width=0.75]    (10.93,-3.29) .. controls (6.95,-1.4) and (3.31,-0.3) .. (0,0) .. controls (3.31,0.3) and (6.95,1.4) .. (10.93,3.29)   ;
\draw   (268.96,142.8) -- (340.96,142.8) -- (340.96,161.27) -- (268.96,161.27) -- cycle ;
\draw    (324.02,142.27) -- (324.02,161.27) ;
\draw    (305.86,142.27) -- (305.86,161.27) ;
\draw    (287.11,142.27) -- (287.11,161.27) ;
\draw    (305.96,181.27) -- (305.96,165.27) ;
\draw [shift={(305.96,163.27)}, rotate = 450] [color={rgb, 255:red, 0; green, 0; blue, 0 }  ][line width=0.75]    (10.93,-3.29) .. controls (6.95,-1.4) and (3.31,-0.3) .. (0,0) .. controls (3.31,0.3) and (6.95,1.4) .. (10.93,3.29)   ;
\draw   (361.96,142.8) -- (433.96,142.8) -- (433.96,161.27) -- (361.96,161.27) -- cycle ;
\draw    (417.02,142.27) -- (417.02,161.27) ;
\draw    (398.86,142.27) -- (398.86,161.27) ;
\draw    (380.11,142.27) -- (380.11,161.27) ;
\draw    (398.96,181.27) -- (398.96,165.27) ;
\draw [shift={(398.96,163.27)}, rotate = 450] [color={rgb, 255:red, 0; green, 0; blue, 0 }  ][line width=0.75]    (10.93,-3.29) .. controls (6.95,-1.4) and (3.31,-0.3) .. (0,0) .. controls (3.31,0.3) and (6.95,1.4) .. (10.93,3.29)   ;
\draw   (174.96,104.8) -- (246.96,104.8) -- (246.96,123.27) -- (174.96,123.27) -- cycle ;
\draw    (230.02,104.27) -- (230.02,123.27) ;
\draw    (211.86,104.27) -- (211.86,123.27) ;
\draw    (193.11,104.27) -- (193.11,123.27) ;
\draw    (211.96,142.27) -- (211.96,126.27) ;
\draw [shift={(211.96,124.27)}, rotate = 450] [color={rgb, 255:red, 0; green, 0; blue, 0 }  ][line width=0.75]    (10.93,-3.29) .. controls (6.95,-1.4) and (3.31,-0.3) .. (0,0) .. controls (3.31,0.3) and (6.95,1.4) .. (10.93,3.29)   ;
\draw   (268.96,103.8) -- (340.96,103.8) -- (340.96,122.27) -- (268.96,122.27) -- cycle ;
\draw    (324.02,103.27) -- (324.02,122.27) ;
\draw    (305.86,103.27) -- (305.86,122.27) ;
\draw    (287.11,103.27) -- (287.11,122.27) ;
\draw    (305.96,142.27) -- (305.96,126.27) ;
\draw [shift={(305.96,124.27)}, rotate = 450] [color={rgb, 255:red, 0; green, 0; blue, 0 }  ][line width=0.75]    (10.93,-3.29) .. controls (6.95,-1.4) and (3.31,-0.3) .. (0,0) .. controls (3.31,0.3) and (6.95,1.4) .. (10.93,3.29)   ;
\draw   (361.96,103.8) -- (433.96,103.8) -- (433.96,122.27) -- (361.96,122.27) -- cycle ;
\draw    (417.02,103.27) -- (417.02,122.27) ;
\draw    (398.86,103.27) -- (398.86,122.27) ;
\draw    (380.11,103.27) -- (380.11,122.27) ;
\draw    (398.96,142.27) -- (398.96,126.27) ;
\draw [shift={(398.96,124.27)}, rotate = 450] [color={rgb, 255:red, 0; green, 0; blue, 0 }  ][line width=0.75]    (10.93,-3.29) .. controls (6.95,-1.4) and (3.31,-0.3) .. (0,0) .. controls (3.31,0.3) and (6.95,1.4) .. (10.93,3.29)   ;
\draw   (439.96,95.27) .. controls (439.96,90.6) and (437.63,88.27) .. (432.96,88.27) -- (409.96,88.27) .. controls (403.29,88.27) and (399.96,85.94) .. (399.96,81.27) .. controls (399.96,85.94) and (396.63,88.27) .. (389.96,88.27)(392.96,88.27) -- (366.96,88.27) .. controls (362.29,88.27) and (359.96,90.6) .. (359.96,95.27) ;
\draw  [color={rgb, 255:red, 155; green, 155; blue, 155 }  ,draw opacity=1 ][dash pattern={on 4.5pt off 4.5pt}] (168.96,99.27) -- (438.96,99.27) -- (438.96,176.27) -- (168.96,176.27) -- cycle ;
\draw   (168.96,181.27) -- (438.96,181.27) -- (438.96,202.27) -- (168.96,202.27) -- cycle ;
\draw    (246.96,113.27) -- (266.96,113.27) ;
\draw [shift={(268.96,113.27)}, rotate = 180] [color={rgb, 255:red, 0; green, 0; blue, 0 }  ][line width=0.75]    (10.93,-3.29) .. controls (6.95,-1.4) and (3.31,-0.3) .. (0,0) .. controls (3.31,0.3) and (6.95,1.4) .. (10.93,3.29)   ;
\draw    (339.96,112.27) -- (359.96,112.27) ;
\draw [shift={(361.96,112.27)}, rotate = 180] [color={rgb, 255:red, 0; green, 0; blue, 0 }  ][line width=0.75]    (10.93,-3.29) .. controls (6.95,-1.4) and (3.31,-0.3) .. (0,0) .. controls (3.31,0.3) and (6.95,1.4) .. (10.93,3.29)   ;
\draw (213,216) node  [align=left] {All};
\draw (306,218) node  [align=left] {birds};
\draw (394,216) node  [align=left] {fly};
\draw (309,191) node  [align=left] {Word Embeddings};
\draw (133,144) node  [align=left] {LSTM};
\draw (401,68) node  [align=left] {Sentence Embedding};
\end{tikzpicture}
}
\caption{Unrolled LSTM encoder }
\end{subfigure}\\
\vspace{10pt}
\begin{subfigure}[b]{0.48\textwidth} 
\scalebox{0.8}{  
\begin{tikzpicture}[x=0.75pt,y=0.75pt,yscale=-1,xscale=1]

\draw  [dash pattern={on 4.5pt off 4.5pt}] (217.96,191) -- (365.96,191) -- (365.96,231) -- (217.96,231) -- cycle ;
\draw    (294.96,253.27) -- (294.96,234.27) ;
\draw [shift={(294.96,232.27)}, rotate = 450] [color={rgb, 255:red, 0; green, 0; blue, 0 }  ][line width=0.75]    (10.93,-3.29) .. controls (6.95,-1.4) and (3.31,-0.3) .. (0,0) .. controls (3.31,0.3) and (6.95,1.4) .. (10.93,3.29)   ;

\draw    (294.96,189.27) -- (294.96,170.27) ;
\draw [shift={(294.96,168.27)}, rotate = 450] [color={rgb, 255:red, 0; green, 0; blue, 0 }  ][line width=0.75]    (10.93,-3.29) .. controls (6.95,-1.4) and (3.31,-0.3) .. (0,0) .. controls (3.31,0.3) and (6.95,1.4) .. (10.93,3.29)   ;

\draw  [dash pattern={on 4.5pt off 4.5pt}] (387.96,191) -- (535.96,191) -- (535.96,231) -- (387.96,231) -- cycle ;
\draw    (464.96,253.27) -- (464.96,234.27) ;
\draw [shift={(464.96,232.27)}, rotate = 450] [color={rgb, 255:red, 0; green, 0; blue, 0 }  ][line width=0.75]    (10.93,-3.29) .. controls (6.95,-1.4) and (3.31,-0.3) .. (0,0) .. controls (3.31,0.3) and (6.95,1.4) .. (10.93,3.29)   ;

\draw    (464.96,189.27) -- (464.96,170.27) ;
\draw [shift={(464.96,168.27)}, rotate = 450] [color={rgb, 255:red, 0; green, 0; blue, 0 }  ][line width=0.75]    (10.93,-3.29) .. controls (6.95,-1.4) and (3.31,-0.3) .. (0,0) .. controls (3.31,0.3) and (6.95,1.4) .. (10.93,3.29)   ;

\draw  [dash pattern={on 4.5pt off 4.5pt}] (272.96,75.27) -- (480.96,75.27) -- (480.96,119.27) -- (272.96,119.27) -- cycle ;
\draw    (293.96,145.27) -- (293.96,121.27) ;
\draw [shift={(293.96,119.27)}, rotate = 450] [color={rgb, 255:red, 0; green, 0; blue, 0 }  ][line width=0.75]    (10.93,-3.29) .. controls (6.95,-1.4) and (3.31,-0.3) .. (0,0) .. controls (3.31,0.3) and (6.95,1.4) .. (10.93,3.29)   ;

\draw    (465.96,147.27) -- (465.96,123.27) ;
\draw [shift={(465.96,121.27)}, rotate = 450] [color={rgb, 255:red, 0; green, 0; blue, 0 }  ][line width=0.75]    (10.93,-3.29) .. controls (6.95,-1.4) and (3.31,-0.3) .. (0,0) .. controls (3.31,0.3) and (6.95,1.4) .. (10.93,3.29)   ;

\draw    (374.96,73.27) -- (374.96,53.27) ;
\draw [shift={(374.96,51.27)}, rotate = 450] [color={rgb, 255:red, 0; green, 0; blue, 0 }  ][line width=0.75]    (10.93,-3.29) .. controls (6.95,-1.4) and (3.31,-0.3) .. (0,0) .. controls (3.31,0.3) and (6.95,1.4) .. (10.93,3.29)   ;

\draw (293,263) node  [align=left] {All birds fly};
\draw (291.98,211) node  [align=left] {Encoder LSTM};
\draw    (217.5,146.5) -- (366.5,146.5) -- (366.5,167.5) -- (217.5,167.5) -- cycle  ;
\draw (292,157) node  [align=left] {Sentence Embedding};
\draw (463,263) node  [align=left] {Penguins fly};
\draw (461.98,211) node  [align=left] {Encoder LSTM};
\draw    (387.5,146.5) -- (536.5,146.5) -- (536.5,167.5) -- (387.5,167.5) -- cycle  ;
\draw (462,157) node  [align=left] {Sentence Embedding};
\draw (376.96,96.27) node  [align=left] {Softmax Inference Classifier};
\draw (377,39) node  [align=left] {\{ Entailment | Contradiction | Neutral \}};

\end{tikzpicture}
}
\caption{InferSent architecture}
\end{subfigure}
\caption[Illustrations of neural sentence embedding architectures based on LSTM encoders.]{Illustrations of neural sentence embedding architectures based on LSTM encoders. (a) shows an unrolled LSTM encoder with word embeddings. (b) shows InferSent architecture with a softmax classification network on top of the encoder. }\label{fig:neur1}
\end{figure}
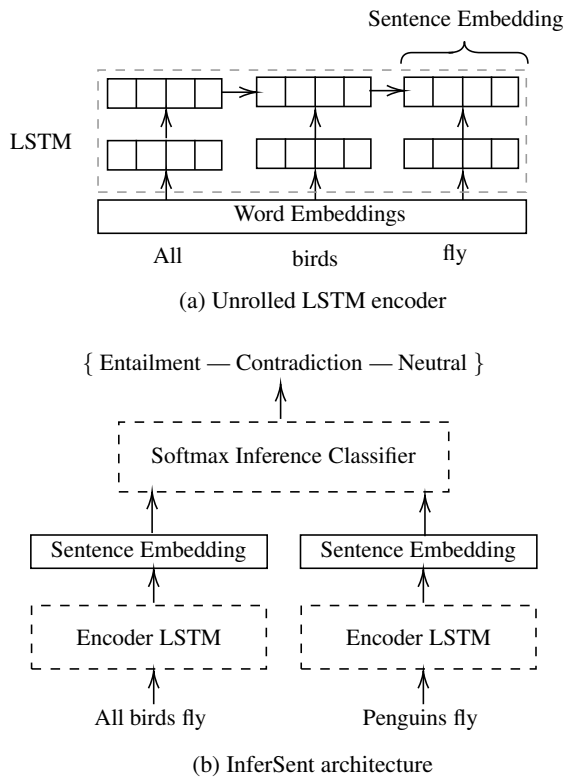

The three objectives above are illustrated in Figures \ref{fig:neur1} and \ref{fig:neur2}. We use the single-layer bidirectional LSTM encoder architecture with max-pooling described in \newcite{conneau2017supervised} for all encoders, and an LSTM decoders for \texttt{SDAE} and \texttt{NMT}. 

\input{tex/neural_figures}

\subsection{Joint Cross-Lingual Modeling}

We first discuss our joint cross-lingual neural models based on the above architectures. Note that joint modeling requires modifying the architecture and objective function of each model in a way that includes simultaneous interactions of cross-lingual sentence embeddings. This can be achieved in various ways with any degree of complexity, but we specifically aim to evaluate a direct extension of each loss function without extraneous objectives or constraints. 

\subsubsection{Joint Cross-Lingual Encoder-Decoder} 
The Sequential Denoising Auto-Encoder (\texttt{SDAE}) is trained to reconstruct the original input sentence from the intermediate sentence representation, where the input is corrupted with linguistic noise, such as word substitutions and reordering \cite{hill2016learning}. This allows the model to robustly learn sentence representations from raw monolingual data. The Neural Machine Translation model, as depicted in Figure \ref{fig:neur2}, has an identical architecture, with the only difference being the language of the input sentence. A cross-lingual extension of \texttt{SDAE} naturally leads to the \texttt{NMT} objective. We combine the \texttt{SDAE} and \texttt{NMT} objectives in a joint architecture, where multiple encoders are trained simultaneously with a single shared decoder. We alternate the input language (and the encoder) in each training batch, and the intermediate sentence embeddings are used as input to the shared decoder. Since the decoder is trained to predict the target sentence from the intermediate sentence representation regardless of input language identity, the encoders are expected to be updated in a way that results in consistent cross-lingual embeddings. Joint multi-lingual NMT has been previously shown to yield cross-lingual representations, as in \newcite{schwenk2017learning}. 

\subsection{Joint Cross-Lingual InferSent}
Since \texttt{InferSent} is trained with an extrinsic classification objective, bilingual or multilingual optimization requires annotated data in each language. At the time of development, the SNLI dataset was only available in English\footnote{Other cross-lingual natural language inference corpora are now publicly available \cite{conneau2018xnli}, but our experiments were conducted before their release.}, so we translated the training and evaluation datasets to Spanish and German using Amazon Translate. Note that in practice, machine translation might not be a viable option, especially if we try to extend the model to low-resource languages. Modern \texttt{NMT} systems require millions of parallel sentences to achieve good translation performance. For our purposes, the translated data allow us to assess the performance in different settings. 

\begin{figure}[h]
\centering       
\scalebox{0.8}{  
\begin{tikzpicture}[x=0.75pt,y=0.75pt,yscale=-1,xscale=1]

\draw  [dash pattern={on 4.5pt off 4.5pt}] (217.96,191) -- (365.96,191) -- (365.96,231) -- (217.96,231) -- cycle ;
\draw    (294.96,253.27) -- (294.96,234.27) ;
\draw [shift={(294.96,232.27)}, rotate = 450] [color={rgb, 255:red, 0; green, 0; blue, 0 }  ][line width=0.75]    (10.93,-3.29) .. controls (6.95,-1.4) and (3.31,-0.3) .. (0,0) .. controls (3.31,0.3) and (6.95,1.4) .. (10.93,3.29)   ;

\draw    (294.96,189.27) -- (294.96,170.27) ;
\draw [shift={(294.96,168.27)}, rotate = 450] [color={rgb, 255:red, 0; green, 0; blue, 0 }  ][line width=0.75]    (10.93,-3.29) .. controls (6.95,-1.4) and (3.31,-0.3) .. (0,0) .. controls (3.31,0.3) and (6.95,1.4) .. (10.93,3.29)   ;

\draw  [dash pattern={on 4.5pt off 4.5pt}] (387.96,191) -- (535.96,191) -- (535.96,231) -- (387.96,231) -- cycle ;
\draw    (464.96,253.27) -- (464.96,234.27) ;
\draw [shift={(464.96,232.27)}, rotate = 450] [color={rgb, 255:red, 0; green, 0; blue, 0 }  ][line width=0.75]    (10.93,-3.29) .. controls (6.95,-1.4) and (3.31,-0.3) .. (0,0) .. controls (3.31,0.3) and (6.95,1.4) .. (10.93,3.29)   ;

\draw    (464.96,189.27) -- (464.96,170.27) ;
\draw [shift={(464.96,168.27)}, rotate = 450] [color={rgb, 255:red, 0; green, 0; blue, 0 }  ][line width=0.75]    (10.93,-3.29) .. controls (6.95,-1.4) and (3.31,-0.3) .. (0,0) .. controls (3.31,0.3) and (6.95,1.4) .. (10.93,3.29)   ;

\draw  [dash pattern={on 4.5pt off 4.5pt}] (272.96,75.27) -- (480.96,75.27) -- (480.96,119.27) -- (272.96,119.27) -- cycle ;
\draw    (293.96,145.27) -- (293.96,121.27) ;
\draw [shift={(293.96,119.27)}, rotate = 450] [color={rgb, 255:red, 0; green, 0; blue, 0 }  ][line width=0.75]    (10.93,-3.29) .. controls (6.95,-1.4) and (3.31,-0.3) .. (0,0) .. controls (3.31,0.3) and (6.95,1.4) .. (10.93,3.29)   ;

\draw    (465.96,147.27) -- (465.96,123.27) ;
\draw [shift={(465.96,121.27)}, rotate = 450] [color={rgb, 255:red, 0; green, 0; blue, 0 }  ][line width=0.75]    (10.93,-3.29) .. controls (6.95,-1.4) and (3.31,-0.3) .. (0,0) .. controls (3.31,0.3) and (6.95,1.4) .. (10.93,3.29)   ;

\draw    (374.96,73.27) -- (374.96,53.27) ;
\draw [shift={(374.96,51.27)}, rotate = 450] [color={rgb, 255:red, 0; green, 0; blue, 0 }  ][line width=0.75]    (10.93,-3.29) .. controls (6.95,-1.4) and (3.31,-0.3) .. (0,0) .. controls (3.31,0.3) and (6.95,1.4) .. (10.93,3.29)   ;

\draw (293,263) node  [align=left] {Alle V{\"o}gel fliegen};
\draw (291.98,211) node  [align=left] {German Encoder};
\draw    (217.5,146.5) -- (366.5,146.5) -- (366.5,167.5) -- (217.5,167.5) -- cycle  ;
\draw (292,157) node  [align=left] {Sentence Embedding};
\draw (463,263) node  [align=left] {Penguins fly};
\draw (461.98,211) node  [align=left] {English Encoder };
\draw    (387.5,146.5) -- (536.5,146.5) -- (536.5,167.5) -- (387.5,167.5) -- cycle  ;
\draw (462,157) node  [align=left] {Sentence Embedding};
\draw (376.96,96.27) node  [align=left] {Softmax Inference Classifier};
\draw (377,39) node  [align=left] {\{ Entailment | Contradiction | Neutral \}};

\end{tikzpicture}
}
\caption{Illustrations of a joint training step, where different languages are used for the premise and hypothesis.}\label{fig:joint}
\end{figure}

\begin{figure*}[h]
\centering
\tikzset{every picture/.style={line width=0.75pt}} 
\scalebox{0.75}{
\begin{tikzpicture}[x=0.75pt,y=0.75pt,yscale=-1,xscale=1]

\draw  [dash pattern={on 4.5pt off 4.5pt}] (124.96,157) -- (272.96,157) -- (272.96,197) -- (124.96,197) -- cycle ;
\draw    (201.96,219.27) -- (201.96,200.27) ;
\draw [shift={(201.96,198.27)}, rotate = 450] [color={rgb, 255:red, 0; green, 0; blue, 0 }  ][line width=0.75]    (10.93,-3.29) .. controls (6.95,-1.4) and (3.31,-0.3) .. (0,0) .. controls (3.31,0.3) and (6.95,1.4) .. (10.93,3.29)   ;

\draw    (201.96,155.27) -- (201.96,136.27) ;
\draw [shift={(201.96,134.27)}, rotate = 450] [color={rgb, 255:red, 0; green, 0; blue, 0 }  ][line width=0.75]    (10.93,-3.29) .. controls (6.95,-1.4) and (3.31,-0.3) .. (0,0) .. controls (3.31,0.3) and (6.95,1.4) .. (10.93,3.29)   ;

\draw  [dash pattern={on 4.5pt off 4.5pt}] (294.96,157) -- (442.96,157) -- (442.96,197) -- (294.96,197) -- cycle ;
\draw    (371.96,219.27) -- (371.96,200.27) ;
\draw [shift={(371.96,198.27)}, rotate = 450] [color={rgb, 255:red, 0; green, 0; blue, 0 }  ][line width=0.75]    (10.93,-3.29) .. controls (6.95,-1.4) and (3.31,-0.3) .. (0,0) .. controls (3.31,0.3) and (6.95,1.4) .. (10.93,3.29)   ;

\draw    (371.96,155.27) -- (371.96,136.27) ;
\draw [shift={(371.96,134.27)}, rotate = 450] [color={rgb, 255:red, 0; green, 0; blue, 0 }  ][line width=0.75]    (10.93,-3.29) .. controls (6.95,-1.4) and (3.31,-0.3) .. (0,0) .. controls (3.31,0.3) and (6.95,1.4) .. (10.93,3.29)   ;

\draw    (200.96,111.27) .. controls (201.5,67.22) and (278.72,-6.26) .. (369.14,61.96) ;
\draw [shift={(370.5,63)}, rotate = 217.57] [color={rgb, 255:red, 0; green, 0; blue, 0 }  ][line width=0.75]    (10.93,-3.29) .. controls (6.95,-1.4) and (3.31,-0.3) .. (0,0) .. controls (3.31,0.3) and (6.95,1.4) .. (10.93,3.29)   ;

\draw   (292.5,65) -- (443.5,65) -- (443.5,88) -- (292.5,88) -- cycle ;
\draw   (445.5,124) .. controls (450.17,124) and (452.5,121.67) .. (452.5,117) -- (452.5,109.5) .. controls (452.5,102.83) and (454.83,99.5) .. (459.5,99.5) .. controls (454.83,99.5) and (452.5,96.17) .. (452.5,89.5)(452.5,92.5) -- (452.5,82) .. controls (452.5,77.33) and (450.17,75) .. (445.5,75) ;
\draw    (486.5,108) -- (486.5,192) ;
\draw [shift={(486.5,194)}, rotate = 270] [color={rgb, 255:red, 0; green, 0; blue, 0 }  ][line width=0.75]    (10.93,-3.29) .. controls (6.95,-1.4) and (3.31,-0.3) .. (0,0) .. controls (3.31,0.3) and (6.95,1.4) .. (10.93,3.29)   ;

\draw   (117.5,149) .. controls (112.83,149) and (110.5,151.33) .. (110.5,156) -- (110.5,169) .. controls (110.5,175.67) and (108.17,179) .. (103.5,179) .. controls (108.17,179) and (110.5,182.33) .. (110.5,189)(110.5,186) -- (110.5,202) .. controls (110.5,206.67) and (112.83,209) .. (117.5,209) ;

\draw (200,229) node  [align=left] {All birds fly};
\draw (198.98,177) node  [align=left] {English Encoder};
\draw    (123.5,112) -- (274.5,112) -- (274.5,134) -- (123.5,134) -- cycle  ;
\draw (199,123) node  [align=left] {Sentence Vector};
\draw (370,229) node  [align=left] {Alle V{\"o}gel fliegen};
\draw (368.98,177) node  [align=left] {German Encoder};
\draw    (293.5,112) -- (444.5,112) -- (444.5,134) -- (293.5,134) -- cycle  ;
\draw (369,123) node  [align=left] {Sentence Vector};
\draw (368,76.5) node  [align=left] {Target Vector};
\draw  [draw opacity=0][fill={rgb, 255:red, 255; green, 255; blue, 255 }  ,fill opacity=1 ]  (224.92,26.96) -- (283.08,26.96) -- (283.08,65.04) -- (224.92,65.04) -- cycle  ;
\draw (254,46) node [scale=0.8,rotate=-336.37] [align=left] {--COPY--};
\draw (491,98) node [scale=0.9] [align=left] {L1 Loss};
\draw  [draw opacity=0][fill={rgb, 255:red, 255; green, 255; blue, 255 }  ,fill opacity=1 ]  (458,134) -- (530,134) -- (530,152) -- (458,152) -- cycle  ;
\draw (494,143) node [scale=0.8] [align=left] {--UPDATE--};
\draw (59,167) node  [align=left] {Pre-trained};
\draw (54,189) node [scale=0.8] [align=left] {--FIXED--};

\end{tikzpicture}
}
\caption{Representation transfer model, with pre-trained English encoder and L1 loss. }\label{fig:trans}
\end{figure*}

Similar to the joint \texttt{SDAE/NMT} model, we train encoders for all languages simultaneously. Since the input to the classifier consists of an ordered pair of sentences, we randomly pick a language for the premise and a language for the hypothesis in each training batch and use their respective encoders. A single classifier is shared regardless of the input languages. Similar to the monolingual case, the model is trained to maximize the performance in the inference classification task, which is cross-lingual in this case. An illustration of a training example is shown in Figure \ref{fig:joint}, where the premise is in German and the hypothesis in English. 

\subsection{Representation Transfer Learning}

In the representation transfer framework, we use a monolingual pre-trained model to guide the training of additional encoders without the original supervised training objective. Using a parallel corpus that has source sentences aligned with English translations, we first generate the representations for the English sentences using a pre-trained \texttt{SDAE} or \texttt{InferSent} model. Then, we use these representations as a target to train an encoder for the other language in a supervised manner. The pivot encoder remains unchanged and only the new encoder is updated during training to ensure that independently trained encoders will still be aligned. Several functions can be used to achieve this, such as the L1 or L2 loss to minimize the distances between the source and target representations, or to maximize the cosine of the angle between them. Empirically, we observed no notable difference between these alternatives.\footnote{We settled on using Adam optimization \cite{kingma2014adam} with L1 loss.} The transfer learning approach is illustrated in Figure \ref{fig:trans}.

\subsection{Sentence Mapping}

We follow the approach used for word-level transformation, where a dictionary is used to fit an orthogonal transformation matrix from the source to the target vector space \cite{smith2017offline}. To extend this to sentences, we use a parallel corpus as a dictionary, and fit a transformation matrix between their sentence embeddings. After training, we apply the learned transformation post-hoc on newly generated sentence embeddings. 

\section{Evaluation}

\input{tex/sdae_figure}

In a well-aligned cross-lingual vector space, sentences should be clustered with their translations across various languages. As discussed in \newcite{schwenk2017learning} this can be measured with sentence translation retrieval: the accuracy of retrieving the correct translation for each source sentence from the target side of a test parallel corpus. This is done using nearest neighbor search with the cosine as a similarity measure. While not exactly an intrinsic evaluation metric, this scheme is the closest measure of alignment quality at the sentence level across all features in the vector space. 

We used bottom-up embeddings composed using weighted averaging with smooth inverse frequency \cite{arora2017asimple,aldarmaki2018evaluation}, which has been shown to work well as monolingual sentence embeddings compared to other bottom-up approaches.  We use skipgram with subword information \cite{bojanowski2017enriching} , i.e. \texttt{FastText}, for the word embeddings, which are also used as input to the neural models. We applied static dictionary alignment using the approach and dictionaries in \newcite{smith2017offline}, in addition to sentence mapping using the parallel corpora. We trained the monolingual \texttt{FastText} word embeddings and SDAE models using the 1 Billion Word benchmark \cite{chelba2014one} for English, and WMT'12 News Crawl data for Spanish and German \cite{wmt-12}. We used WMT'12 Common Crawl data for cross-lingual alignment, and WMT'12 test sets for evaluations. We used the augmented SNLI data described in \cite{dasgupta2018evaluating} and their translations for training the mono-lingual and joint \texttt{InferSent} models. For all datasets and languages, the only preprocessing performed was tokenization.

One of our evaluation objective is to assess the minimal bilingual data requirements for each framework, so we split the training parallel corpora into subsets of increasing size from 1,000 to 1 million sentences, where we double the size in each step. We report sentence translation retrieval accuracies in all language directions, using \textit{en} for English, \textit{es} for Spanish, and \textit{de} for German \footnote{This evaluation scheme was recently introduced in \newcite{aldarmaki2019context} with data splits that are now available for download. Note that we used slightly older datasets in our experiments.}.

\subsection{Results}

\input{tex/infer_figure}

The results of the various \texttt{SDAE} models compared with the baselines are shown in Figure \ref{fig:nn_acc_sdae}. With less than 100K parallel sentences, the joint \texttt{SDAE/NMT} model yielded poor performance compared to all models, but with 100K and more data, the model quickly exceeded the performance of all others by a large margin. Transfer learning achieved the second best performance, although it lagged behind the joint model with large parallel sets. With small amounts of parallel text, all models outperformed the joint \texttt{SDAE/NMT}, particularly the word based \texttt{FastText} models. Sentence mapping performed on average better than the static dictionary baseline, but \texttt{FastText} sentence mapping was generally better. 

Figure \ref{fig:nn_acc_inf} shows the results of the \texttt{InferSent} alignment models. Note that the joint InferSent model was trained with supervision using the translated SNLI data instead of the variable-size parallel corpora, so the performance is constant with respect to the number of parallel sentences. The joint model did not learn to align the cross-lingual sentences. Possible explanations of this failure are discussed in section \ref{sec:infer}. 

Overall, the transfer learning model worked well for \texttt{InferSent}  resulting in high translation retrieval accuracies even with relatively small amounts of parallel text ($\sim$ 5K sentences). Sentence mapping also performed better than the word-based baselines with additional parallel data ($>$ 20K).

\subsection{Overall Evaluation}

In this section, we compare the overall performance of different types of models on sentence translation retrieval. We plotted the average cross-lingual accuracy (averaged over all language directions) by the best performing variant of each model in Figure \ref{fig:overall}. With small amounts of parallel text, around 5K sentences, the best performance was achieved using \texttt{InferSent} transfer model. The model continued to yield the highest performance until it was exceeded by the joint \texttt{SDAE/NMT} model at 500K sentences. The representation transfer models for \texttt{SDAE} exceeded the \texttt{FastText} model at around 20K sentences, and achieved comparable performance to \texttt{InferSent} sentence mapping.

\begin{figure}
\begin{subfigure}{0.49\textwidth}
\hspace{-30pt}
\centering
\small
 \begin{tikzpicture}[font=\sffamily, scale=0.8, transform shape]
\begin{semilogyaxis}[title=Average ,
                ymode=normal,
                 title style={at={(0.5,1.0)},anchor=north,yshift=-0.1},
                ymin=0,
                ymax=100,
                xmin=0,
                 ytick={20, 50, 80, 100},
                yticklabels={.2, .5, .8, 1},
                xtick={2, 5, 20, 200, 500},
                xticklabels={2K, 5K, 20K, 200K, .5M}, 
                xmode=log, 
                log basis x={10},
                axis x line=bottom,
                axis y line=left,
                axis line style=-,
                x=2.5cm, y=0.05cm,
                compat=newest,
                ylabel = NN accuracy,
                xlabel = \# parallel sentences,
                 legend style={at={(1,-0.4)},anchor= east,draw=white!15!black,legend cell align=left},
                ]

\addplot[magenta, mark=x,mark options={solid, scale=1, fill=black}] table [x=size, y=sdae_joint, col sep=space] {data/overall_avg.nn};
\addplot[brown, dotted, mark=x,mark options={solid, scale=1, fill=black}] table [x=size, y=sdae_trans, col sep=space] {data/overall_avg.nn};
\addplot[black, mark=*,mark options={solid, scale=1, fill=black}] table [x=size, y=Infer_trans, col sep=space] {data/overall_avg.nn};
\addplot[blue, dotted, mark=*,mark options={solid, scale=0.7}] table [x=size, y=infer_sent, col sep=space] {data/overall_avg.nn};
\addplot[green,mark=none,mark options={solid, scale=1}] table [x=size, y=sif_sent, col sep=space] {data/overall_avg.nn};
\addplot [dashed] table [x=size, y=sif_word, col sep=space] {data/overall_avg.nn};
\draw [dotted] (2,\pgfkeysvalueof{/pgfplots/ymin}) -- (2,90);
\draw [dotted] (20,\pgfkeysvalueof{/pgfplots/ymin}) -- (20,90);
\draw [dotted] (200,\pgfkeysvalueof{/pgfplots/ymin}) -- (200,98);

\addlegendimage{magenta,, mark=x,mark options={solid, scale=1, fill=black}} 
\addlegendentry{SDAE/NMT \texttt{(joint)} 93.42\%};
\addlegendimage{brown, dotted, mark=x,mark options={solid, scale=1, fill=black}} 
\addlegendentry{SDAE \texttt{(Transfer)} 69.18\%};
\addlegendimage{black, mark=*,mark options={solid, scale=1, fill=black}} 
\addlegendentry{InferSent \texttt{(Transfer)} 85.99\%};
\addlegendimage{blue, dotted, mark=*,mark options={solid, scale=0.7},legend cell align=left}
\addlegendentry{InferSent \texttt{(sent)} 76.76\%};
\addlegendimage{green,mark=none,mark options={solid, scale=1}}
\addlegendentry{FastText \texttt{(sent)} 69.18\%};
\addlegendimage {dashed}
\addlegendentry{FastText \texttt{(dict)} 57.45\%};
\end{semilogyaxis}
\end{tikzpicture}
\end{subfigure}
\caption[Nearest neighbor translation accuracy of various models ]{Nearest neighbor translation accuracy as a function of (log) parallel corpus size. The legend shows the average accuracies of each model using 1M parallel sentences.} \label{fig:overall}
\end{figure}

\subsection{Analysis of Joint InferSent Performance}\label{sec:infer}

 \begin{table}[t]
 \centering
    \scalebox{0.8}{
   \hspace{-10pt}
       \setlength{\tabcolsep}{2.5pt}
    \begin{tabular}{|m{1.8cm}|m{7.5cm}|}
      \multicolumn{1} {c}{ \textbf{Language}} & \multicolumn{1} {c} {\textbf{Monolingual Nearest Neighbors}}  \\
      \hline
         \multicolumn{2} {l}{Query:  Tons of people are gathered around the statue } \\  
         \hline
         \hline
             \multirow{1}{1.5cm}{\centering Spanish} & There are \textbf{several people sitting around} a table. \newline There are \textbf{several people} outside of a building. \newline \textbf{There are multiple people} present.  \\
          \cline{2-2}
         \multirow{1}{1.7cm}{\centering English} &  The \textbf{people} are taking photos of \textbf{the statue}.\newline  \textbf{A group of people} looking at \textbf{a statue}. \newline \textbf{People are gathered} by the water. \\
         \hline
         \multicolumn{2} {l}{Query: A vehicle is crossing a river} \\  
         \hline
         \hline 
            \multirow{1}{1.5cm}{\centering Spanish} & \textbf{A sedan} is stuck in the middle of \textbf{a river}.\newline People are \textbf{crossing a river}. \newline A \textbf{taxi cab} is \textbf{driving down a path of snow}. \\
             \cline{2-2}
             \multirow{1}{1.7cm}{\centering English} & A person is near \textbf{a river}.\newline People are \textbf{crossing a river}. \newline A \textbf{Land Rover} is splashing water as it \textbf{crosses a river}. \\
             \hline
      \end{tabular}
      }
           \caption[Monolingual Nearest neighbors using joint InferSent encoders.]{Mono-lingual nearest neighbors (or their translations) of a sample of query sentences from SNLI test set using joint \texttt{InferSent} encoders. Phrases similar to the query sentences are shown in \textbf{bold}.}\label{tab:inf1}
           \end{table}

 \begin{table}[h]
 \centering
    \scalebox{0.8}{
   \hspace{-10pt}
       \setlength{\tabcolsep}{2.5pt}
    \begin{tabular}{|m{1.8cm}|m{7.5cm}|}
       \multicolumn{1} {c}{\textbf{Language}} &  \multicolumn{1} {c} {\textbf{Cross-lingual Nearest Neighbors}} \\
      \hline
         \multicolumn{2} {l}{Query: Tons of people are gathered around the statue } \\  
         \hline
         \hline
             \multirow{1}{1.5cm}{\centering Spanish} & Food and wine are on the table that has \textbf{many people surrounding} it.\newline \textbf{Some people} enjoying their brunch together in the outdoor seating area of a restaurant...\newline The \textbf{group of people} are game developers creating a new video game in their office. \\
          \cline{2-2}
         \multirow{1}{1.7cm}{\centering English} &  The \textbf{group of people} are flying in the air on their unicorns .\newline A \textbf{group of people} are standing around with smiles on their faces... \newline A \textbf{group of people} dressed as clowns stroll into the Bigtop Circus holding signs. \\
         \hline
         \multicolumn{2} {l}{Query: A vehicle is crossing a river} \\  
         \hline
         \hline 
            \multirow{1}{1.5cm}{\centering Spanish}  & People and a baby are \textbf{crossing} the street at a crosswalk to get home.\newline The person in the picture is riding \textbf{a bike slowing up hill} , pumping the pedals as hard as they can. \newline The man , wearing scuba gear , jumps off the side of \textbf{the boat into the ocean} below.\\
             \cline{2-2}
             \multirow{1}{1.7cm}{\centering English} & A person in a coat with a briefcase \textbf{walks down a street} next to the bus lane.\newline A man waterskiing in \textbf{a river} with a large wall in the background.\newline A person \textbf{waterskiing in a river} with a wall in the background.\\
             \hline
      \end{tabular}
      }
           \caption[Cross-lingual Nearest neighbors using joint InferSent encoders.]{Cross-lingual nearest neighbors (or their translations) of a sample of query sentences from SNLI test set using joint \texttt{InferSent} encoders. Phrases similar to the query sentences are shown in \textbf{bold}.}\label{tab:inf}
           \end{table}
 
The joint \texttt{InferSent} model was trained to maximize the cross-lingual classification accuracy on cross-lingual inference data. The cross-lingual inference classification performance was comparable to the monolingual case for each language. The monolingual accuracies were around 83\%, 79\%, and 79\% for English, German, and Spanish, respectively. The cross-lingual accuracy was around 79\%.  Given this relatively high performance in NLI classification and the poor performance in cross-lingual translation retrieval, we surmise that the 3-way classification objective is not demanding enough to learn general-purpose semantic representations. In addition, high performance in a specific extrinsic evaluation task is not necessarily an indication of general embedding quality. 
           
Tables \ref{tab:inf1} and \ref{tab:inf} show examples of monolingual and cross-lingual nearest neighbors (or their English translations) from the hypotheses in SNLI test sets.  The cross-lingual nearest neighbors did share several semantic aspects with the query sentence; subjects or verbs or combinations of these were observed in nearest neighbors. However, the exact translations were not the nearest neighbors in most cases, and the nearest neighbors often included several extraneous pieces of content not present in the query sentence. The mono-lingual nearest neighbors, on the other hand, were more semantically similar to each other, not only in the semantic features that are present, but also in their exclusions of dissimilar details. 

We surmise that only a subset of semantic features were learned by the \texttt{InferSent} objective given the specific characteristics of the SNLI training sets. In other words, the model was not pushed to preserve the full semantic content since only a small subset of features were useful for entailment relationships. The higher similarity among monolingual nearest neighbors is likely an artifact of the underlying word embeddings passing through the same encoder network.

\subsection{Extrinsic Evaluation}

Relying on a single measure is never sufficient to probe all characteristics of a vector space. 
Extrinsic evaluation can be another useful tool to measure the effectiveness of various cross-lingual models, although extrinsic tasks typically measure specific and narrow aspects of semantics. Nevertheless, we can still gain some insights about certain characteristics of these models and their applicability. One of the most widely used tasks for cross-lingual evaluation is the Cross-Lingual Document Classification benchmark (CLDC), where a model is trained in one language and tested on another \cite{schwenk2018corpus,Klement-12}.

We report the average classification accuracies in CLDC across all language directions (a total of six directions) using the datasets in \newcite{schwenk2018corpus}; the multi-layer perceptron was used as a classifier trained for each source language, then tested in the remaining two. 

The highest accuracy was achieved using \texttt{FastText} vectors, followed by \texttt{InferSent} transfer and sentence mapping models. With large enough parallel corpora, the performance of \texttt{SDAE/NMT} exceeded the transfer model, but with smaller data, \texttt{SDAE} transfer model achieved consistently higher performance. 

These results are consistent with the trend of these models in mono-lingual topic categorization \cite{aldarmaki2018evaluation}, where word averaging achieved consistently higher performance than all neural models. This indicates that cross-lingual models share the same semantic characteristics as their underlying mono-lingual counterparts. We should underscore that CLDC is a rather coarse categorization task where documents are classified into four categories. Note also that the \texttt{FastText} model achieved relatively high performance even when it was aligned with only 1K parallel sentences, a condition in which sentence translation retrieval accuracy was less that 40\%. This poor correlation with sentence translation retrieval accuracies indicates that neither evaluation framework is reliable on its own. Our intuition is that sentence translation retrieval is a more comprehensive measure since all features in the vector space weigh equally in calculating the cosine similarity; on the other hand, a supervised classifier weighs features according to their correlations with the target classes.

\begin{figure}
\begin{subfigure}{0.49\textwidth}
\centering
\small
\hspace{-30pt}
 \begin{tikzpicture}[font=\sffamily, scale=0.8, transform shape]
\begin{semilogyaxis}[title=Average ,
                ymode=normal,
                 title style={at={(0.5,1.0)},anchor=north,yshift=-0.1},
                ymin=0,
                ymax=100,
                xmin=0,
                 ytick={20, 50, 80, 100},
                yticklabels={.2, .5, .8, 1},
                xtick={2, 5, 20, 200, 500},
                xticklabels={2K, 5K, 20K, 200K, .5M}, 
                xmode=log, 
                log basis x={10},
                axis x line=bottom,
                axis y line=left,
                axis line style=-,
                x=2.5cm, y=0.05cm,
                ylabel = CLDC accuracy,
                compat=newest,
                xlabel = \# parallel sentences,
               legend style={at={(1,-0.4)},anchor= east,draw=white!15!black,legend cell align=left}
                ]

\addplot[magenta, mark=x,mark options={solid, scale=1, fill=black}] table [x=size, y=sdae_nmt, col sep=space] {data/extrinsic_avg.nn};
\addplot[brown, dotted, mark=x,mark options={solid, scale=1, fill=black}] table [x=size, y=sdae_trans, col sep=space] {data/extrinsic_avg.nn};
\addplot[black, mark=*,mark options={solid, scale=1, fill=black}] table [x=size, y=infer, col sep=space] {data/extrinsic_avg.nn};
\addplot[blue, dotted, mark=*,mark options={solid, scale=0.7}] table [x=size, y=infer_sent, col sep=space] {data/extrinsic_avg.nn};
\addplot[red, dotted, mark=x,mark options={solid, scale=1, fill=black}] table [x=size, y=infer_joint, col sep=space] {data/extrinsic_avg.nn};
\addplot[green,mark=none,mark options={solid, scale=1}] table [x=size, y=siskip_sent, col sep=space] {data/extrinsic_avg.nn};
\addlegendimage{magenta,, mark=x,mark options={solid, scale=1, fill=black}} 
\addlegendentry{SDAE/NMT \texttt{(joint)} 73.26\%};
\addlegendimage{brown, dotted, mark=x,mark options={solid, scale=1, fill=black}} 
\addlegendentry{SDAE \texttt{(Transfer)} 71.48\%};
\addlegendimage{black, mark=*,mark options={solid, scale=1, fill=black}} 
\addlegendentry{InferSent \texttt{(Transfer)} 77.43\%};
\addlegendimage{blue, dotted, mark=*,mark options={solid, scale=0.7}}
\addlegendentry{InferSent \texttt{(sent)} 76.14\%};
\addlegendimage{red, dotted, mark=x,mark options={solid, scale=1, fill=black}} 
\addlegendentry{InferSent \texttt{(joint)} 37.86\%};
\addlegendimage{green,mark=none,mark options={solid, scale=1}}
\addlegendentry{FastText \texttt{(sent)} 81.84\%};
\end{semilogyaxis}
\end{tikzpicture}

\end{subfigure}
\caption[Average cross-lingual document classification accuracy]{Average cross-lingual document classification accuracy as a function of (log) parallel corpus size. The legend shows the average accuracies of each model using 1M parallel sentences.} \label{fig:nn_acc}
\end{figure}

\section{Conclusions}

We explored different approaches for cross-lingual alignment of top-down sentence embedding models: joint modeling, representation transfer, and sentence mapping. 
With sufficient amounts of parallel text, joint modeling yielded superior performance in the joint SDAE and NMT model, while joint InferSent failed to yield good alignments. Our results underscore the difficulty of joint modeling itself in addition to its relatively high data and memory requirements. With smaller amounts of parallel text, representation transfer worked reasonably well across all models, whereas sentence mapping was generally worse. Moreover, the transfer and sentence mapping frameworks enable modular training where additional languages can be added without retraining existing models and without labeled training data (as in InferSent), which allows scaling neural models to more languages with less resources. In extrinsic evaluation using cross-lingual document classification, transfer models achieved consistently better performance than joint models. Between the two sentence embedding models we evaluated, InferSent yielded better performance than SDAE and NMT, except in the joint framework. 

In practice, joint and transfer learning can be combined in various ways according to data availability and modeling choices. A multi-task framework can be used to optimize both objectives at once. Given the lower data cost of representation transfer models, a joint model can be trained first for a set of resource-rich languages, followed by transfer learning for low-resource languages.

\bibliography{refs}
\bibliographystyle{acl_natbib}

\end{document}

%% file: tex/neural_figures.tex
\begin{figure}[H]
\centering
\begin{subfigure}[b]{0.24\textwidth} 
\scalebox{0.8}{  
\begin{tikzpicture}[x=0.75pt,y=0.75pt,yscale=-1,xscale=1]
\draw  [dash pattern={on 4.5pt off 4.5pt}] (217.96,191) -- (365.96,191) -- (365.96,231) -- (217.96,231) -- cycle ;
\draw    (294.96,253.27) -- (294.96,234.27) ;
\draw [shift={(294.96,232.27)}, rotate = 450] [color={rgb, 255:red, 0; green, 0; blue, 0 }  ][line width=0.75]    (10.93,-3.29) .. controls (6.95,-1.4) and (3.31,-0.3) .. (0,0) .. controls (3.31,0.3) and (6.95,1.4) .. (10.93,3.29)   ;

\draw    (294.96,189.27) -- (294.96,170.27) ;
\draw [shift={(294.96,168.27)}, rotate = 450] [color={rgb, 255:red, 0; green, 0; blue, 0 }  ][line width=0.75]    (10.93,-3.29) .. controls (6.95,-1.4) and (3.31,-0.3) .. (0,0) .. controls (3.31,0.3) and (6.95,1.4) .. (10.93,3.29)   ;

\draw  [dash pattern={on 4.5pt off 4.5pt}] (217.96,85) -- (365.96,85) -- (365.96,125) -- (217.96,125) -- cycle ;
\draw    (294.96,146.27) -- (294.96,127.27) ;
\draw [shift={(294.96,125.27)}, rotate = 450] [color={rgb, 255:red, 0; green, 0; blue, 0 }  ][line width=0.75]    (10.93,-3.29) .. controls (6.95,-1.4) and (3.31,-0.3) .. (0,0) .. controls (3.31,0.3) and (6.95,1.4) .. (10.93,3.29)   ;

\draw    (294.96,81.27) -- (294.96,61.27) ;
\draw [shift={(294.96,59.27)}, rotate = 450] [color={rgb, 255:red, 0; green, 0; blue, 0 }  ][line width=0.75]    (10.93,-3.29) .. controls (6.95,-1.4) and (3.31,-0.3) .. (0,0) .. controls (3.31,0.3) and (6.95,1.4) .. (10.93,3.29)   ;

\draw (293,263) node  [align=left] {All birds fly};
\draw (291.98,211) node  [align=left] {Encoder LSTM};
\draw    (217.5,146.5) -- (366.5,146.5) -- (366.5,167.5) -- (217.5,167.5) -- cycle  ;
\draw (292,157) node  [align=left] {Sentence Embedding};
\draw (291.98,105) node  [align=left] {Decoder LSTM};
\draw (296,46) node  [align=left] {All birds fly};
\end{tikzpicture}
}
\caption[]{SAE objective}
\end{subfigure}
\begin{subfigure}[b]{0.23\textwidth} 
\scalebox{0.8}{  
\begin{tikzpicture}[x=0.75pt,y=0.75pt,yscale=-1,xscale=1]

\draw  [dash pattern={on 4.5pt off 4.5pt}] (217.96,191) -- (365.96,191) -- (365.96,231) -- (217.96,231) -- cycle ;
\draw    (294.96,253.27) -- (294.96,234.27) ;
\draw [shift={(294.96,232.27)}, rotate = 450] [color={rgb, 255:red, 0; green, 0; blue, 0 }  ][line width=0.75]    (10.93,-3.29) .. controls (6.95,-1.4) and (3.31,-0.3) .. (0,0) .. controls (3.31,0.3) and (6.95,1.4) .. (10.93,3.29)   ;

\draw    (294.96,189.27) -- (294.96,170.27) ;
\draw [shift={(294.96,168.27)}, rotate = 450] [color={rgb, 255:red, 0; green, 0; blue, 0 }  ][line width=0.75]    (10.93,-3.29) .. controls (6.95,-1.4) and (3.31,-0.3) .. (0,0) .. controls (3.31,0.3) and (6.95,1.4) .. (10.93,3.29)   ;

\draw  [dash pattern={on 4.5pt off 4.5pt}] (217.96,85) -- (365.96,85) -- (365.96,125) -- (217.96,125) -- cycle ;
\draw    (294.96,146.27) -- (294.96,127.27) ;
\draw [shift={(294.96,125.27)}, rotate = 450] [color={rgb, 255:red, 0; green, 0; blue, 0 }  ][line width=0.75]    (10.93,-3.29) .. controls (6.95,-1.4) and (3.31,-0.3) .. (0,0) .. controls (3.31,0.3) and (6.95,1.4) .. (10.93,3.29)   ;

\draw    (294.96,81.27) -- (294.96,61.27) ;
\draw [shift={(294.96,59.27)}, rotate = 450] [color={rgb, 255:red, 0; green, 0; blue, 0 }  ][line width=0.75]    (10.93,-3.29) .. controls (6.95,-1.4) and (3.31,-0.3) .. (0,0) .. controls (3.31,0.3) and (6.95,1.4) .. (10.93,3.29)   ;

\draw (293,263) node  [align=left] {Alle V{\"o}gel fliegen};
\draw (291.98,211) node  [align=left] {Encoder LSTM};
\draw    (217.5,146.5) -- (366.5,146.5) -- (366.5,167.5) -- (217.5,167.5) -- cycle  ;
\draw (292,157) node  [align=left] {Sentence Embedding};
\draw (291.98,105) node  [align=left] {Decoder LSTM};
\draw (296,46) node  [align=left] {All birds fly};

\end{tikzpicture}
}
\caption[]{NMT objective}
\end{subfigure}
\caption[Illustrations of LSTM encoder-decoder architectures for sentence embeddings]{Illustrations of LSTM encoder-decoder architectures for sentence embeddings. (a) Sequential Auto-Encoder objective, where the input and output are the same sentence. (b) Neural Machine Translation objective, where the output is a translation of the input sentence from a parallel corpus.}\label{fig:neur2}
\end{figure}

%% file: tex/sdae_figure.tex
\begin{figure}[t]
\hspace{-0.2cm}
\begin{subfigure}{0.24\textwidth}
\centering
\small
 \begin{tikzpicture}[font=\sffamily, scale=0.80, transform shape]
\begin{semilogyaxis}[title=es $\rightarrow$ en ,
                ymode=normal,
                ymin=0,
                ymax=100,
                xmin=0,
                ytick={20, 50, 80, 100},
                yticklabels={.2, .5, .8, 1},
                xtick={1, 10,100, 1000},
                xticklabels={1K,10K,100K, 1M}, 
                xmode=log, 
                log basis x={2},
                axis x line=bottom,
                axis y line=left,
                axis line style=-,
                x=0.35cm, y=0.03cm,
                ylabel = NN accuracy,
                compat=newest,
                y label style={at={(-0.1,0.5)}},
                ]
\addplot[blue, dotted, mark=*,mark options={solid, scale=0.7}] table [x=size, y=elmo_avg, col sep=space] {data/sdae_es_en.nn};
\addplot[gray,mark=o,mark options={solid, scale=0.7}] table [x=size, y=elmo_sent, col sep=space] {data/sdae_es_en.nn};
\addplot[magenta, dotted, mark=x,mark options={solid, scale=1, fill=black}] table [x=size, y=siskip_sif, col sep=space] {data/sdae_es_en.nn};
\addplot[green,mark=none,mark options={solid, scale=1}] table [x=size, y=siskip_sent, col sep=space] {data/sdae_es_en.nn};
\addplot [dashed] table [x=size, y=muse_sif, col sep=space] {data/sdae_es_en.nn};
\end{semilogyaxis}
\end{tikzpicture}

\end{subfigure}
\begin{subfigure}{0.24\textwidth}
\centering
\small
 \begin{tikzpicture}[font=\sffamily, scale=0.80, transform shape]
\begin{semilogyaxis}[title=en $\rightarrow$ es ,
                ymode=normal,
                ymin=0,
                ymax=100,
                xmin=0,
                 ytick={20, 50, 80, 100},
                yticklabels={.2, .5, .8, 1},
                xtick={1, 10,100, 1000},
                xticklabels={1K,10K,100K, 1M}, 
                xmode=log, 
                log basis x={2},
                axis x line=bottom,
                axis y line=left,
                axis line style=-,
                x=0.35cm, y=0.03cm,
                ]
\addplot[blue, dotted, mark=*,mark options={solid, scale=0.7}] table [x=size, y=elmo_avg, col sep=space] {data/sdae_en_es.nn};
\addplot[gray,mark=o,mark options={solid, scale=0.7}] table [x=size, y=elmo_sent, col sep=space] {data/sdae_en_es.nn};
\addplot[magenta, dotted, mark=x,mark options={solid, scale=1, fill=black}] table [x=size, y=siskip_sif, col sep=space] {data/sdae_en_es.nn};
\addplot[green,mark=none,mark options={solid, scale=1}] table [x=size, y=siskip_sent, col sep=space] {data/sdae_en_es.nn};
\addplot [dashed] table [x=size, y=muse_sif, col sep=space] {data/sdae_en_es.nn};
\end{semilogyaxis}
\end{tikzpicture}

\end{subfigure}\\
\begin{subfigure}{0.24\textwidth}
\centering
\small
 \begin{tikzpicture}[font=\sffamily, scale=0.80, transform shape]
\begin{semilogyaxis}[title=de $\rightarrow$ en ,
                 title style={at={(0.5,1.0)},anchor=north,yshift=-0.1},
                ymode=normal,
                ymin=0,
                ymax=100,
                xmin=0,
                 ytick={20, 50, 80, 100},
                yticklabels={.2, .5, .8, 1},
                xtick={1, 10,100, 1000},
                xticklabels={1K,10K,100K, 1M}, 
                xmode=log, 
                log basis x={2},
                axis x line=bottom,
                axis y line=left,
                axis line style=-,
                x=0.35cm, y=0.03cm,
                compat=newest,
                ylabel = NN accuracy,
                y label style={at={(-0.1,0.5)}},
                                ]

\addplot[blue, dotted, mark=*,mark options={solid, scale=0.7}] table [x=size, y=elmo_avg, col sep=space] {data/sdae_de_en.nn};
\addplot[gray,mark=o,mark options={solid, scale=0.7}] table [x=size, y=elmo_sent, col sep=space] {data/sdae_de_en.nn};
\addplot[magenta, dotted, mark=x,mark options={solid, scale=1, fill=black}] table [x=size, y=siskip_sif, col sep=space] {data/sdae_de_en.nn};
\addplot[green,mark=none,mark options={solid, scale=1}] table [x=size, y=siskip_sent, col sep=space] {data/sdae_de_en.nn};
\addplot [dashed] table [x=size, y=muse_sif, col sep=space] {data/sdae_de_en.nn};
\end{semilogyaxis}
\end{tikzpicture}

\end{subfigure}%
\begin{subfigure}{0.24\textwidth}
\centering
\small
 \begin{tikzpicture}[font=\sffamily, scale=0.80, transform shape]
\begin{semilogyaxis}[title=en $\rightarrow$ de ,
                title style={at={(0.5,1.0)},anchor=north,yshift=-0.1},
                ymode=normal,
                ymin=0,
                ymax=100,
                xmin=0,
                 ytick={20, 50, 80, 100},
                yticklabels={.2, .5, .8, 1},
                xtick={1, 10,100, 1000},
                xticklabels={1K,10K,100K, 1M}, 
                xmode=log, 
                log basis x={2},
                axis x line=bottom,
                axis y line=left,
                axis line style=-,
                x=0.35cm, y=0.03cm,
                compat=newest,
                y label style={at={(-0.1,0.5)}},
                ]
\addplot[blue, dotted, mark=*,mark options={solid, scale=0.7}] table [x=size, y=elmo_avg, col sep=space] {data/sdae_en_de.nn};
\addplot[gray,mark=o,mark options={solid, scale=0.7}] table [x=size, y=elmo_sent, col sep=space] {data/sdae_en_de.nn};
\addplot[magenta, dotted, mark=x,mark options={solid, scale=1, fill=black}] table [x=size, y=siskip_sif, col sep=space] {data/sdae_en_de.nn};
\addplot[green,mark=none,mark options={solid, scale=1}] table [x=size, y=siskip_sent, col sep=space] {data/sdae_en_de.nn};
\addplot [dashed] table [x=size, y=muse_sif, col sep=space] {data/sdae_en_de.nn};
\end{semilogyaxis}
\end{tikzpicture}

\end{subfigure}\\
\begin{subfigure}{0.24\textwidth}
\centering
\small
\hspace{-10pt}
 \begin{tikzpicture}[font=\sffamily, scale=0.80, transform shape]
\begin{semilogyaxis}[title=es $\rightarrow$ de ,
                ymode=normal,
                 title style={at={(0.5,1.0)},anchor=north,yshift=-0.1},
                ymin=0,
                ymax=100,
                xmin=1,
                 ytick={20, 50, 80, 100},
                yticklabels={.2, .5, .8, 1},
                xtick={1, 10,100, 1000},
                xticklabels={1K,10K,100K, 1M}, 
                xmode=log, 
                log basis x={2},
                axis x line=bottom,
                axis y line=left,
                axis line style=-,
                x=0.35cm, y=0.03cm,
                compat=newest,
                ylabel = NN accuracy,
                xlabel = \# parallel sentences
                ]
\addplot[blue, dotted, mark=*,mark options={solid, scale=0.7}] table [x=size, y=elmo_avg, col sep=space] {data/sdae_en_de.nn};
\addplot[gray,mark=o,mark options={solid, scale=0.7}] table [x=size, y=elmo_sent, col sep=space] {data/sdae_en_de.nn};
\addplot[magenta, dotted, mark=x,mark options={solid, scale=1, fill=black}] table [x=size, y=siskip_sif, col sep=space] {data/sdae_en_de.nn};
\addplot[green,mark=none,mark options={solid, scale=1}] table [x=size, y=siskip_sent, col sep=space] {data/sdae_en_de.nn};
\addplot [dashed] table [x=size, y=muse_sif, col sep=space] {data/sdae_en_de.nn};
\end{semilogyaxis}

\end{tikzpicture}
\end{subfigure}
\begin{subfigure}{0.24\textwidth}
\centering
\small
 \begin{tikzpicture}[font=\sffamily, scale=0.80, transform shape]
\begin{semilogyaxis}[title=de $\rightarrow$ es ,
                ymode=normal,
                 title style={at={(0.5,1.0)},anchor=north,yshift=-0.1},
                ymin=0,
                ymax=100,
                xmin=0,
                 ytick={20, 50, 80, 100},
                yticklabels={.2, .5, .8, 1},
                xtick={1, 10,100, 1000},
                xticklabels={1K,10K,100K, 1M}, 
                xmode=log, 
                log basis x={2},
                axis x line=bottom,
                axis y line=left,
                axis line style=-,
                x=0.35cm, y=0.03cm,
                xlabel = \# parallel sentences
                ]
\addplot[blue, dotted, mark=*,mark options={solid, scale=0.7}] table [x=size, y=elmo_avg, col sep=space] {data/sdae_de_es.nn};
\addplot[gray,mark=o,mark options={solid, scale=0.7}] table [x=size, y=elmo_sent, col sep=space] {data/sdae_de_es.nn};
\addplot[magenta, dotted, mark=x,mark options={solid, scale=1, fill=black}] table [x=size, y=siskip_sif, col sep=space] {data/sdae_de_es.nn};
\addplot[green,mark=none,mark options={solid, scale=1}] table [x=size, y=siskip_sent, col sep=space] {data/sdae_de_es.nn};
\addplot [dashed] table [x=size, y=muse_sif, col sep=space] {data/sdae_de_es.nn};
\end{semilogyaxis}
\end{tikzpicture}
\end{subfigure}\\
\begin{subfigure}{0.5\textwidth}
\small
\centering
 \begin{tikzpicture}[font=\sffamily, scale=0.750, transform shape]
 \centering
\begin{axis}[
    hide axis,
    xmin=0,
    xmax=0.1,
    ymin=0,
    ymax=10,
      compat=newest,
    legend style={at={(1.3,0.8)},anchor=east,draw=white!15!black,legend cell align=left},    
    ]
\addlegendimage{blue, dotted, mark=*,mark options={solid, scale=0.7}}
\addlegendentry{SDAE \texttt{(transfer)} 75.92\%};
\addlegendimage{gray,mark=o,mark options={solid, scale=0.7}}
\addlegendentry{SDAE \texttt{(sent)} 67.57\%};
\addlegendimage{magenta, dotted, mark=x,mark options={solid, scale=1, fill=black}} 
\addlegendentry{SDAE/NMT \texttt{(joint)} 93.42\%};
\addlegendimage{green,mark=none,mark options={solid, scale=1}}
\addlegendentry{FastText \texttt{(sent)} 69.18\%};
\addlegendimage {dashed}
\addlegendentry{FastText \texttt{(dict)} 57.45\%};
\end{axis}
\end{tikzpicture}
\end{subfigure}
\caption[Nearest neighbor translation accuracy for SDAE models ]{Nearest neighbor translation accuracy as a function of (log) parallel corpus size. \texttt{(sent)} to sentence-level mapping, and \texttt{(dict)} refers to the baseline (using a static dictionary for mapping). The legend shows the average accuracies of each model using 1M parallel sentences.} \label{fig:nn_acc_sdae}
\end{figure}

%% file: tex/infer_figure.tex
\begin{figure}[t]
\hspace{-0.2cm}
\begin{subfigure}{0.24\textwidth}
\centering
\small
 \begin{tikzpicture}[font=\sffamily, scale=0.80, transform shape]
\begin{semilogyaxis}[title=es $\rightarrow$ en ,
                ymode=normal,
                ymin=0,
                ymax=100,
                xmin=0,
                ytick={20, 50, 80, 100},
                yticklabels={.2, .5, .8, 1},
                xtick={1, 10,100, 1000},
                xticklabels={1K,10K,100K, 1M}, 
                xmode=log, 
                log basis x={2},
                axis x line=bottom,
                axis y line=left,
                axis line style=-,
                x=0.35cm, y=0.03cm,
                ylabel = NN accuracy,
                compat=newest,
                y label style={at={(-0.1,0.5)}},
                ]
\addplot[black, dotted, mark=*,mark options={solid, scale=0.7}] table [x=size, y=elmo_avg, col sep=space] {data/infer_es_en.nn};
\addplot[gray,mark=o,mark options={solid, scale=0.7}] table [x=size, y=elmo_sent, col sep=space] {data/infer_es_en.nn};
\addplot[red, dotted, mark=x,mark options={solid, scale=1, fill=black}] table [x=size, y=siskip_sif, col sep=space] {data/infer_es_en.nn};
\addplot[green,mark=none,mark options={solid, scale=1}] table [x=size, y=siskip_sent, col sep=space] {data/infer_es_en.nn};
\addplot [dashed] table [x=size, y=muse_sif, col sep=space] {data/infer_es_en.nn};
\end{semilogyaxis}
\end{tikzpicture}

\end{subfigure}
\begin{subfigure}{0.23\textwidth}
\centering
\small
 \begin{tikzpicture}[font=\sffamily, scale=0.80, transform shape]
\begin{semilogyaxis}[title=en $\rightarrow$ es ,
                ymode=normal,
                ymin=0,
                ymax=100,
                xmin=0,
                 ytick={20, 50, 80, 100},
                yticklabels={.2, .5, .8, 1},
                xtick={1, 10,100, 1000},
                xticklabels={1K,10K,100K, 1M}, 
                xmode=log, 
                log basis x={2},
                axis x line=bottom,
                axis y line=left,
                axis line style=-,
                x=0.35cm, y=0.03cm,
                compat=newest,
                y label style={at={(-0.1,0.5)}},
                ]
\addplot[black, dotted, mark=*,mark options={solid, scale=0.7}] table [x=size, y=elmo_avg, col sep=space] {data/infer_en_es.nn};
\addplot[gray,mark=o,mark options={solid, scale=0.7}] table [x=size, y=elmo_sent, col sep=space] {data/infer_en_es.nn};
\addplot[red, dotted, mark=x,mark options={solid, scale=1, fill=black}] table [x=size, y=siskip_sif, col sep=space] {data/infer_en_es.nn};
\addplot[green,mark=none,mark options={solid, scale=1}] table [x=size, y=siskip_sent, col sep=space] {data/infer_en_es.nn};
\addplot [dashed] table [x=size, y=muse_sif, col sep=space] {data/infer_en_es.nn};
\end{semilogyaxis}
\end{tikzpicture}
\end{subfigure}\\
\hspace{-0.2cm}
\begin{subfigure}{0.24\textwidth}
\centering
\small
 \begin{tikzpicture}[font=\sffamily, scale=0.80, transform shape]
\begin{semilogyaxis}[title=de $\rightarrow$ en ,
                ymode=normal,
                 title style={at={(0.5,1.0)},anchor=north,yshift=-0.1},
                ymin=0,
                ymax=100,
                xmin=1,
                 ytick={20, 50, 80, 100},
                yticklabels={.2, .5, .8, 1},
                xtick={1, 10,100, 1000},
                xticklabels={1K,10K,100K, 1M}, 
                xmode=log, 
                log basis x={2},
                axis x line=bottom,
                axis y line=left,
                axis line style=-,
                x=0.35cm, y=0.03cm,
                ylabel = NN accuracy,
                compat=newest,
                y label style={at={(-0.1,0.5)}},
                ]
\addplot[black, dotted, mark=*,mark options={solid, scale=0.7}] table [x=size, y=elmo_avg, col sep=space] {data/infer_de_en.nn};
\addplot[gray,mark=o,mark options={solid, scale=0.7}] table [x=size, y=elmo_sent, col sep=space] {data/infer_de_en.nn};
\addplot[red, dotted, mark=x,mark options={solid, scale=1, fill=black}] table [x=size, y=siskip_sif, col sep=space] {data/infer_de_en.nn};
\addplot[green,mark=none,mark options={solid, scale=1}] table [x=size, y=siskip_sent, col sep=space] {data/infer_de_en.nn};
\addplot [dashed] table [x=size, y=muse_sif, col sep=space] {data/infer_de_en.nn};
\end{semilogyaxis}
\end{tikzpicture}

\end{subfigure}
\begin{subfigure}{0.23\textwidth}
\centering
\small
 \begin{tikzpicture}[font=\sffamily, scale=0.80, transform shape]
\begin{semilogyaxis}[title=en $\rightarrow$ de ,
                title style={at={(0.5,1.0)},anchor=north,yshift=-0.1},
                ymode=normal,
                ymin=0,
                ymax=100,
                xmin=0,
                 ytick={20, 50, 80, 100},
                yticklabels={.2, .5, .8, 1},
                xtick={1, 10,100, 1000},
                xticklabels={1K,10K,100K, 1M}, 
                xmode=log, 
                log basis x={2},
                axis x line=bottom,
                axis y line=left,
                axis line style=-,
                x=0.35cm, y=0.03cm,
                ]
\addplot[black, dotted, mark=*,mark options={solid, scale=0.7}] table [x=size, y=elmo_avg, col sep=space] {data/infer_en_de.nn};
\addplot[gray,mark=o,mark options={solid, scale=0.7}] table [x=size, y=elmo_sent, col sep=space] {data/infer_en_de.nn};
\addplot[red, dotted, mark=x,mark options={solid, scale=1, fill=black}] table [x=size, y=siskip_sif, col sep=space] {data/infer_en_de.nn};
\addplot[green,mark=none,mark options={solid, scale=1}] table [x=size, y=siskip_sent, col sep=space] {data/infer_en_de.nn};
\addplot [dashed] table [x=size, y=muse_sif, col sep=space] {data/infer_en_de.nn};
\end{semilogyaxis}
\end{tikzpicture}

\end{subfigure}\\
\begin{subfigure}{0.24\textwidth}
\centering
\small
 \begin{tikzpicture}[font=\sffamily, scale=0.80, transform shape]
\begin{semilogyaxis}[title=es $\rightarrow$ de ,
                title style={at={(0.5,1.0)},anchor=north,yshift=-0.1},
                ymode=normal,
                ymin=0,
                ymax=100,
                xmin=0,
                 ytick={20, 50, 80, 100},
                yticklabels={.2, .5, .8, 1},
                xtick={1, 10,100, 1000},
                xticklabels={1K,10K,100K, 1M}, 
                xmode=log, 
                log basis x={2},
                axis x line=bottom,
                axis y line=left,
                axis line style=-,
                x=0.35cm, y=0.03cm,
                ylabel = NN accuracy,
                compat=newest,
                xlabel = \# parallel sentences,
                y label style={at={(-0.1,0.5)}},
                                ]

\addplot[black, dotted, mark=*,mark options={solid, scale=0.7}] table [x=size, y=elmo_avg, col sep=space] {data/infer_es_de.nn};
\addplot[gray,mark=o,mark options={solid, scale=0.7}] table [x=size, y=elmo_sent, col sep=space] {data/infer_es_de.nn};
\addplot[red, dotted, mark=x,mark options={solid, scale=1, fill=black}] table [x=size, y=siskip_sif, col sep=space] {data/infer_es_de.nn};
\addplot[green,mark=none,mark options={solid, scale=1}] table [x=size, y=siskip_sent, col sep=space] {data/infer_es_de.nn};
\addplot [dashed] table [x=size, y=muse_sif, col sep=space] {data/infer_es_de.nn};
\end{semilogyaxis}
\end{tikzpicture}
\end{subfigure}
\begin{subfigure}{0.23\textwidth}
\centering
\small
 \begin{tikzpicture}[font=\sffamily, scale=0.80, transform shape]
\begin{semilogyaxis}[title=de $\rightarrow$ es ,
                ymode=normal,
                title style={at={(0.5,1.0)},anchor=north,yshift=-0.1},
                ymin=0,
                ymax=100,
                xmin=0,
                 ytick={20, 50, 80, 100},
                yticklabels={.2, .5, .8, 1},
                xtick={1, 10,100, 1000},
                xticklabels={1K,10K,100K, 1M}, 
                xmode=log, 
                log basis x={2},
                axis x line=bottom,
                axis y line=left,
                axis line style=-,
                x=0.35cm, y=0.03cm,
                xlabel = \# parallel sentences
                ]
\addplot[black, dotted, mark=*,mark options={solid, scale=0.7}] table [x=size, y=elmo_avg, col sep=space] {data/infer_de_es.nn};
\addplot[gray,mark=o,mark options={solid, scale=0.7}] table [x=size, y=elmo_sent, col sep=space] {data/infer_de_es.nn};
\addplot[red, dotted, mark=x,mark options={solid, scale=1, fill=black}] table [x=size, y=siskip_sif, col sep=space] {data/infer_de_es.nn};
\addplot[green,mark=none,mark options={solid, scale=1}] table [x=size, y=siskip_sent, col sep=space] {data/infer_de_es.nn};
\addplot [dashed] table [x=size, y=muse_sif, col sep=space] {data/infer_de_es.nn};
\end{semilogyaxis}
\end{tikzpicture}
\end{subfigure}\\
\begin{subfigure}{0.5\textwidth}
\small
\centering
 \begin{tikzpicture}[font=\sffamily, scale=0.750, transform shape]
 \centering
\begin{axis}[
    hide axis,
    xmin=0,
    xmax=0.5,
    ymin=0,
    ymax=0.4,
     compat=newest,
    legend style={at={(1.3,0.8)},anchor=east,draw=white!15!black,legend cell align=left},
    ]
\addlegendimage{black, dotted, mark=*,mark options={solid, scale=0.7}}
\addlegendentry{InferSent \texttt{(transfer)} 85.99\%};
\addlegendimage{gray,mark=o,mark options={solid, scale=0.7}}
\addlegendentry{InferSent \texttt{(sent)} 76.76\%};
\addlegendimage{red, dotted, mark=x,mark options={solid, scale=1, fill=black}} 
\addlegendentry{InferSent \texttt{(joint)} 18.88\%};
\addlegendimage{green,mark=none,mark options={solid, scale=1}}
\addlegendentry{FastText \texttt{(sent)} 69.18\%};
\addlegendimage {dashed}
\addlegendentry{FastText \texttt{(dict)} 57.45\%};
\end{axis}
\end{tikzpicture}
\end{subfigure}
\caption[Nearest neighbor translation accuracy of InferSent models]{Nearest neighbor translation accuracy as a function of (log) parallel corpus size. \texttt{(sent)} to sentence-level mapping, and \texttt{(dict)} refers to the baseline (using a static dictionary for mapping). The legend shows the average accuracies of each model using 1M parallel sentences.} \label{fig:nn_acc_inf}
\end{figure}

%% file: naaclhlt2019.bbl
\begin{thebibliography}{33}
\expandafter\ifx\csname natexlab\endcsname\relax\def\natexlab#1{#1}\fi

\bibitem[{Aldarmaki and Diab(2016)}]{aldarmaki2016learning}
Hanan Aldarmaki and Mona Diab. 2016.
\newblock Learning cross-lingual representations with matrix factorization.
\newblock In \emph{Proceedings of the Workshop on Multilingual and
  Cross-lingual Methods in NLP}, pages 1--9.

\bibitem[{Aldarmaki and Diab(2018)}]{aldarmaki2018evaluation}
Hanan Aldarmaki and Mona Diab. 2018.
\newblock Evaluation of unsupervised compositional representations.
\newblock \emph{Proceedings of the 27th International Conference on
  Computational Linguistics}.

\bibitem[{Aldarmaki and Diab(2019)}]{aldarmaki2019context}
Hanan Aldarmaki and Mona Diab. 2019.
\newblock Context-aware crosslingual mapping.
\newblock \emph{arXiv preprint arXiv:1903.03243}.

\bibitem[{Aldarmaki et~al.(2018)Aldarmaki, Mohan, and
  Diab}]{aldarmaki2018unsupervised}
Hanan Aldarmaki, Mahesh Mohan, and Mona Diab. 2018.
\newblock Unsupervised word mapping using structural similarities in
  monolingual embeddings.
\newblock \emph{Transactions of the Association of Computational Linguistics},
  6.

\bibitem[{Ammar et~al.(2016)Ammar, Mulcaire, Tsvetkov, Lample, Dyer, and
  Smith}]{AmmarMTLDS16}
Waleed Ammar, George Mulcaire, Yulia Tsvetkov, Guillaume Lample, Chris Dyer,
  and Noah~A Smith. 2016.
\newblock Massively multilingual word embeddings.
\newblock \emph{arXiv preprint arXiv:1602.01925}.

\bibitem[{Arora et~al.(2017)Arora, Liang, and Ma}]{arora2017asimple}
Sanjeev Arora, Yingyu Liang, and Tengyu Ma. 2017.
\newblock A simple but tough-to-beat baseline for sentence embeddings.

\bibitem[{Bojanowski et~al.(2017)Bojanowski, Grave, Joulin, and
  Mikolov}]{bojanowski2017enriching}
Piotr Bojanowski, Edouard Grave, Armand Joulin, and Tomas Mikolov. 2017.
\newblock Enriching word vectors with subword information.
\newblock \emph{Transactions of the Association for Computational Linguistics},
  5:135--146.

\bibitem[{Bowman et~al.(2015sss)Bowman, Angeli, Potts, and
  Manning}]{bowmanlarge}
Samuel~R Bowman, Gabor Angeli, Christopher Potts, and Christopher~D Manning.
  2015sss.
\newblock A large annotated corpus for learning natural language inference.
\newblock \emph{Proceedings of the 2015 Conference on Empirical Methods in
  Natural Language Processing}.

\bibitem[{Callison-Burch et~al.(2012)Callison-Burch, Koehn, Monz, Post,
  Soricut, and Specia}]{wmt-12}
Chris Callison-Burch, Philipp Koehn, Christof Monz, Matt Post, Radu Soricut,
  and Lucia Specia. 2012.
\newblock Findings of the 2012 workshop on statistical machine translation.
\newblock In \emph{Proceedings of the Seventh Workshop on Statistical Machine
  Translation}, pages 10--51.

\bibitem[{Chelba et~al.(2014)Chelba, Mikolov, Schuster, Ge, Brants, Koehn, and
  Robinson}]{chelba2014one}
Ciprian Chelba, Tomas Mikolov, Mike Schuster, Qi~Ge, Thorsten Brants, Phillipp
  Koehn, and Tony Robinson. 2014.
\newblock One billion word benchmark for measuring progress in statistical
  language modeling.
\newblock In \emph{Fifteenth Annual Conference of the International Speech
  Communication Association}.

\bibitem[{Conneau and Kiela(2018)}]{conneau2018senteval}
Alexis Conneau and Douwe Kiela. 2018.
\newblock Senteval: An evaluation toolkit for universal sentence
  representations.
\newblock In \emph{Proceedings of the Eleventh International Conference on
  Language Resources and Evaluation (LREC-2018)}.

\bibitem[{Conneau et~al.(2017{\natexlab{a}})Conneau, Kiela, Schwenk, Barrault,
  and Bordes}]{conneau2017supervised}
Alexis Conneau, Douwe Kiela, Holger Schwenk, Lo{\"\i}c Barrault, and Antoine
  Bordes. 2017{\natexlab{a}}.
\newblock Supervised learning of universal sentence representations from
  natural language inference data.
\newblock In \emph{Proceedings of the 2017 Conference on Empirical Methods in
  Natural Language Processing}, pages 670--680.

\bibitem[{Conneau et~al.(2017{\natexlab{b}})Conneau, Lample, Ranzato, Denoyer,
  and J{\'e}gou}]{conneau2017word}
Alexis Conneau, Guillaume Lample, Marc'Aurelio Ranzato, Ludovic Denoyer, and
  Herv{\'e} J{\'e}gou. 2017{\natexlab{b}}.
\newblock Word translation without parallel data.
\newblock \emph{arXiv preprint arXiv:1710.04087}.

\bibitem[{Conneau et~al.(2018)Conneau, Rinott, Lample, Williams, Bowman,
  Schwenk, and Stoyanov}]{conneau2018xnli}
Alexis Conneau, Ruty Rinott, Guillaume Lample, Adina Williams, Samuel~R.
  Bowman, Holger Schwenk, and Veselin Stoyanov. 2018.
\newblock Xnli: Evaluating cross-lingual sentence representations.
\newblock In \emph{Proceedings of the 2018 Conference on Empirical Methods in
  Natural Language Processing}. Association for Computational Linguistics.

\bibitem[{Dasgupta et~al.(2018)Dasgupta, Guo, Stuhlm{\"u}ller, Gershman, and
  Goodman}]{dasgupta2018evaluating}
Ishita Dasgupta, Demi Guo, Andreas Stuhlm{\"u}ller, Samuel~J Gershman, and
  Noah~D Goodman. 2018.
\newblock Evaluating compositionality in sentence embeddings.
\newblock \emph{arXiv preprint arXiv:1802.04302}.

\bibitem[{Gouws et~al.(2015)Gouws, Bengio, and Corrado}]{gouws2015bilbowa}
Stephan Gouws, Yoshua Bengio, and Greg Corrado. 2015.
\newblock Bilbowa: Fast bilingual distributed representations without word
  alignments.
\newblock In \emph{Proceedings of the 32nd International Conference on Machine
  Learning (ICML-15)}, pages 748--756.

\bibitem[{Guo et~al.(2018)Guo, Shen, Yang, Ge, Cer, Abrego, Stevens, Constant,
  Sung, Strope et~al.}]{guo2018effective}
Mandy Guo, Qinlan Shen, Yinfei Yang, Heming Ge, Daniel Cer, Gustavo~Hernandez
  Abrego, Keith Stevens, Noah Constant, Yun-hsuan Sung, Brian Strope, et~al.
  2018.
\newblock Effective parallel corpus mining using bilingual sentence embeddings.
\newblock In \emph{Proceedings of the Third Conference on Machine Translation:
  Research Papers}, pages 165--176.

\bibitem[{Hill et~al.(2016)Hill, Cho, and Korhonen}]{hill2016learning}
Felix Hill, Kyunghyun Cho, and Anna Korhonen. 2016.
\newblock Learning distributed representations of sentences from unlabelled
  data.
\newblock In \emph{Proceedings of NAACL-HLT}, pages 1367--1377.

\bibitem[{Johnson et~al.(2017)Johnson, Schuster, Le, Krikun, Wu, Chen, Thorat,
  Vi{\'e}gas, Wattenberg, Corrado et~al.}]{johnson2016google}
Melvin Johnson, Mike Schuster, Quoc~V Le, Maxim Krikun, Yonghui Wu, Zhifeng
  Chen, Nikhil Thorat, Fernanda Vi{\'e}gas, Martin Wattenberg, Greg Corrado,
  et~al. 2017.
\newblock Google's multilingual neural machine translation system: Enabling
  zero-shot translation.
\newblock \emph{Transactions of the Association of Computational Linguistics},
  5(1):339--351.

\bibitem[{Kingma and Ba(2014)}]{kingma2014adam}
Diederik~P Kingma and Jimmy Ba. 2014.
\newblock Adam: A method for stochastic optimization.
\newblock \emph{arXiv preprint arXiv:1412.6980}.

\bibitem[{Kiros et~al.(2015)Kiros, Zhu, Salakhutdinov, Zemel, Urtasun,
  Torralba, and Fidler}]{kiros2015skip}
Ryan Kiros, Yukun Zhu, Ruslan~R Salakhutdinov, Richard Zemel, Raquel Urtasun,
  Antonio Torralba, and Sanja Fidler. 2015.
\newblock Skip-thought vectors.
\newblock In \emph{Advances in neural information processing systems}, pages
  3294--3302.

\bibitem[{Klementiev et~al.(2012)Klementiev, Titov, and Bhattarai}]{Klement-12}
Alexandre Klementiev, Ivan Titov, and Binod Bhattarai. 2012.
\newblock Inducing crosslingual distributed representations of words.
\newblock \emph{Proceedings of COLING 2012}, pages 1459--1474.

\bibitem[{Le and Mikolov(2014)}]{le2014distributed}
Quoc Le and Tomas Mikolov. 2014.
\newblock Distributed representations of sentences and documents.
\newblock In \emph{International Conference on Machine Learning}, pages
  1188--1196.

\bibitem[{Pham et~al.(2015)Pham, Luong, and Manning}]{Pham-15}
Hieu Pham, Thang Luong, and Christopher Manning. 2015.
\newblock Learning distributed representations for multilingual text sequences.
\newblock In \emph{Proceedings of the 1st Workshop on Vector Space Modeling for
  Natural Language Processing}, pages 88--94.

\bibitem[{Schuster and Paliwal(1997)}]{schuster1997bidirectional}
Mike Schuster and Kuldip~K Paliwal. 1997.
\newblock Bidirectional recurrent neural networks.
\newblock \emph{IEEE Transactions on Signal Processing}, 45(11):2673--2681.

\bibitem[{Schwenk and Douze(2017)}]{schwenk2017learning}
Holger Schwenk and Matthijs Douze. 2017.
\newblock Learning joint multilingual sentence representations with neural
  machine translation.
\newblock In \emph{Proceedings of the 2nd Workshop on Representation Learning
  for NLP}, pages 157--167.

\bibitem[{Schwenk and Li(2018)}]{schwenk2018corpus}
Holger Schwenk and Xian Li. 2018.
\newblock A corpus for multilingual document classification in eight languages.
\newblock In \emph{Proceedings of the Eleventh International Conference on
  Language Resources and Evaluation (LREC-2018)}.

\bibitem[{Shi et~al.(2015)Shi, Liu, Liu, and Sun}]{glove-15}
Tianze Shi, Zhiyuan Liu, Yang Liu, and Maosong Sun. 2015.
\newblock Learning cross-lingual word embeddings via matrix co-factorization.
\newblock In \emph{Proceedings of the 53rd Annual Meeting of the Association
  for Computational Linguistics and the 7th International Joint Conference on
  Natural Language Processing (Volume 2: Short Papers)}, volume~2, pages
  567--572.

\bibitem[{Smith et~al.(2017)Smith, Turban, Hamblin, and
  Hammerla}]{smith2017offline}
Samuel~L Smith, David~HP Turban, Steven Hamblin, and Nils~Y Hammerla. 2017.
\newblock Offline bilingual word vectors, orthogonal transformations and the
  inverted softmax.
\newblock \emph{arXiv preprint arXiv:1702.03859}.

\bibitem[{Soyer et~al.(2014)Soyer, Stenetorp, and Aizawa}]{soyer2014leveraging}
Hubert Soyer, Pontus Stenetorp, and Akiko Aizawa. 2014.
\newblock Leveraging monolingual data for crosslingual compositional word
  representations.
\newblock \emph{arXiv preprint arXiv:1412.6334}.

\bibitem[{Sutskever et~al.(2014)Sutskever, Vinyals, and
  Le}]{sutskever2014sequence}
Ilya Sutskever, Oriol Vinyals, and Quoc~V Le. 2014.
\newblock Sequence to sequence learning with neural networks.
\newblock In \emph{Advances in neural information processing systems}, pages
  3104--3112.

\bibitem[{Upadhyay et~al.(2016)Upadhyay, Faruqui, Dyer, and
  Roth}]{upadhyay2016cross}
Shyam Upadhyay, Manaal Faruqui, Chris Dyer, and Dan Roth. 2016.
\newblock Cross-lingual models of word embeddings: An empirical comparison.
\newblock In \emph{Proceedings of the 54th Annual Meeting of the Association
  for Computational Linguistics (Volume 1: Long Papers)}, volume~1.

\bibitem[{Zhou et~al.(2016)Zhou, Wan, and Xiao}]{zhou2016cross}
Xinjie Zhou, Xiaojun Wan, and Jianguo Xiao. 2016.
\newblock Cross-lingual sentiment classification with bilingual document
  representation learning.
\newblock In \emph{Proceedings of the 54th Annual Meeting of the Association
  for Computational Linguistics (Volume 1: Long Papers)}, volume~1, pages
  1403--1412.

\end{thebibliography}
